\documentclass[1p]{elsarticle}
\usepackage{amssymb}
\usepackage{multirow}
\usepackage{synttree}
\usepackage{lineno,hyperref}

\usepackage[utf8]{inputenc}
\usepackage[T2A]{fontenc}
\usepackage{float}
\usepackage{color,soul}

\usepackage[table,dvipsnames]{xcolor}

\modulolinenumbers[5]
\usepackage{amsmath}
\usepackage{url}
\usepackage{color}
\usepackage{enumitem}
\usepackage{xcolor}
\usepackage{float}
\definecolor{darkgreen}{rgb}{0.0, 0.5, 0.0}
\definecolor{Gray}{gray}{0.9}
\definecolor{GrayD}{gray}{0.4}
\definecolor{LightCyan}{rgb}{0.88,1,1}

\usepackage{array}   

\usepackage{algorithmic}
\usepackage{graphicx}
\usepackage{subcaption}

\usepackage{comment}
\usepackage{todonotes}
\usepackage{booktabs}
\usepackage{xltabular}
\usepackage{hyperref}
\usepackage{dcolumn}
\usepackage[per-mode=symbol]{siunitx} 

\usepackage[normalem]{ulem}
    
\journal{Artificial Intelligence in Medicine}

\biboptions{authoryear}

\usepackage{listings}
\usepackage{color}

\definecolor{dkgreen}{rgb}{0,0.6,0}
\definecolor{gray}{rgb}{0.5,0.5,0.5}
\definecolor{mauve}{rgb}{0.58,0,0.82}
\definecolor{gray}{rgb}{0.4,0.4,0.4}
\definecolor{darkblue}{rgb}{0.0,0.0,0.6}
\definecolor{lightblue}{rgb}{0.0,0.0,0.9}
\definecolor{cyan}{rgb}{0.0,0.6,0.6}
\definecolor{darkred}{rgb}{0.6,0.0,0.0}
\definecolor{lightgray}{rgb}{0.97,0.97,0.97}

\lstdefinelanguage{XML}
{
  morestring=[s][\color{mauve}]{"}{"},
  morestring=[s][\color{black}]{>}{<},
  morecomment=[s]{<?}{?>},
  morecomment=[s][\color{dkgreen}]{<!--}{-->},
  stringstyle=\color{black},
  identifierstyle=\color{lightblue},
  keywordstyle=\color{red},
  morekeywords={xmlns,xsi,noNamespaceSchemaLocation,type,id,x,y,source,target,version,tool,transRef,roleRef,objective,eventually}
}

\begin{document}

\begin{frontmatter}

\title{MedExpQA: Multilingual Benchmarking of Large Language Models for Medical Question Answering}

\author[mymainaddress]{I\~nigo Alonso}
\ead{inigoborja.alonso@ehu.eus}

\author[mymainaddress]{Maite Oronoz}
\ead{maite.oronoz@ehu.eus}

\author[mymainaddress]{Rodrigo Agerri\corref{mycorrespondingauthor}}
\ead{rodrigo.agerri@ehu.eus}

\cortext[mycorrespondingauthor]{Corresponding author}

\address[mymainaddress]{HiTZ Center - Ixa, University of the Basque Country UPV/EHU}

\begin{abstract}
Large Language Models (LLMs) have the potential of facilitating the development of Artificial Intelligence technology to assist medical experts for interactive decision support. This potential has been illustrated by the state-of-the-art performance obtained by LLMs in Medical Question Answering, with striking results such as passing marks in licensing medical exams. However, while impressive, the required quality bar for medical applications remains far from being achieved. Currently, LLMs remain challenged by outdated knowledge and by their tendency to generate hallucinated content. Furthermore, most benchmarks to assess medical knowledge lack reference gold explanations which means that it is not possible to evaluate the reasoning of LLMs predictions. Finally, the situation is particularly grim if we consider benchmarking LLMs for languages other than English which remains, as far as we know, a totally neglected topic. \textcolor{black}{In order to address these shortcomings, in this paper we present MedExpQA, the first multilingual benchmark based on medical exams to evaluate LLMs in Medical Question Answering. To the best of our knowledge, MedExpQA includes for the first time reference gold explanations, written by medical doctors, of the correct and incorrect options in the exams. Comprehensive multilingual experimentation using both the gold reference explanations and Retrieval Augmented Generation (RAG) approaches show that performance of LLMs, with best results around 75 accuracy for English, still has large room for improvement, especially for languages other than English, for which accuracy drops 10 points. Therefore, despite using state-of-the-art RAG methods, our results also demonstrate the difficulty of obtaining and integrating readily available medical knowledge that may positively impact results on downstream evaluations for Medical Question Answering.} 
Data, code, and fine-tuned models will be made publicly available\footnote{\url{https://huggingface.co/datasets/HiTZ/MedExpQA}}

\end{abstract}

\begin{keyword}
Large Language Models\sep Medical Question Answering \sep Multilinguality \sep Retrieval Augmented Generation \sep Natural Language Processing
\end{keyword}

\end{frontmatter}


\section{Introduction}\label{sec:introduction}

We are currently seeing a dramatic increase in research on how to apply Artificial Intelligence (AI) to the medical domain with the aim of generating decision support tools to assist medical experts in their everyday activities. This has been further motivated by rather strong claims about Large Language Models (LLMs) in medical Question Answering (QA) tasks, such as that they obtain passing marks for medical licensing exams like the United States Medical Licensing Examination (USMLE) \citep{singhal2023large,nori2023capabilities}.

Assisting medical experts by answering their medical questions is a natural way of articulating human-AI interaction as it is usually considered that Medical QA involves processing, acquiring and summarizing relevant information and knowledge and then reasoning about how to apply the available knowledge to the current context given by a clinical case. For example, a resident medical doctor preparing for the licensing exams may want to know what and why is the correct treatment or diagnosis in the context of a clinical case \citep{LLMMedEduc:23,goenaga2023explanatory}. This means that a LLM should be able to automatically identify, access and correctly apply the relevant medical knowledge, and that it will be capable of elucidating between the variety of symptoms, each of which may be indicative of multiple diseases. Finally, it is also assumed that the model will interact with the resident medical doctor in a natural manner, ideally using natural language. Therefore, developing the required AI technology to help, for example, resident medical doctors to prepare their licensing exams remains a far from trivial endeavour.

Nonetheless, and as a crucial first step to address this challenge, the AI ecosystem has seen an explosion of LLMs (both general purpose and specific to the medical domain) reporting high accuracy results on Medical QA tasks thereby demonstrating that LLMs are somewhat capable of encoding clinical knowledge \citep{singhal2023large}. State-of-the-art models include publicly available ones such as LLaMA \citep{touvron2023llama} and the medical-specific PMC-LLaMA \citep{wu2023pmcllama}, Mistral \citep{jiang2023mistral} and its medical version BioMistral \citep{labrak2024biomistral}, and proprietary models such as MedPaLM \citep{singhal2023towards} and GPT-4 \citep{nori2023capabilities}, among many others.

While their published high-accuracy scores on Medical QA may seem impressive, these LLMs still present a number of shortcomings. First, LLMs usually generate factually inaccurate answers that seem plausible enough for a non-medical expert (known as hallucinations) \citep{xie2023faithful,xiong2024benchmarking}. Second, their knowledge might be outdated as the pre-training data used to train the LLMs may not include the latest available medical knowledge. Third, the Medical QA benchmarks \citep{singhal2023large,xiong2024benchmarking} on which they are evaluated do not include gold reference explanations generated by medical doctors that provide the required reasoning to support the model's predictions. Finally, and to the best of our knowledge, evaluations have only been done for English, which makes it impossible to know how well these LLMs fare for other languages.

Retrieval Augmented Generation (RAG) techniques have been specifically proposed to address the first two issues, namely, the lack of up-to-date medical knowledge and the tendency of these models to hallucinate \citep{xiong2024benchmarking}. Their \textsc{MedRAG} approach obtains clear zero-shot improvements for two of the five datasets on their \textsc{MIRAGE} benchmark, while for the rest the obtained gains are rather modest. Still, \textsc{MedRAG} proves to be an effective technique to improve Medical QA by incorporating external medical knowledge \citep{xiong2024benchmarking}.

\begin{figure}
\centering
\includegraphics[width=\textwidth]{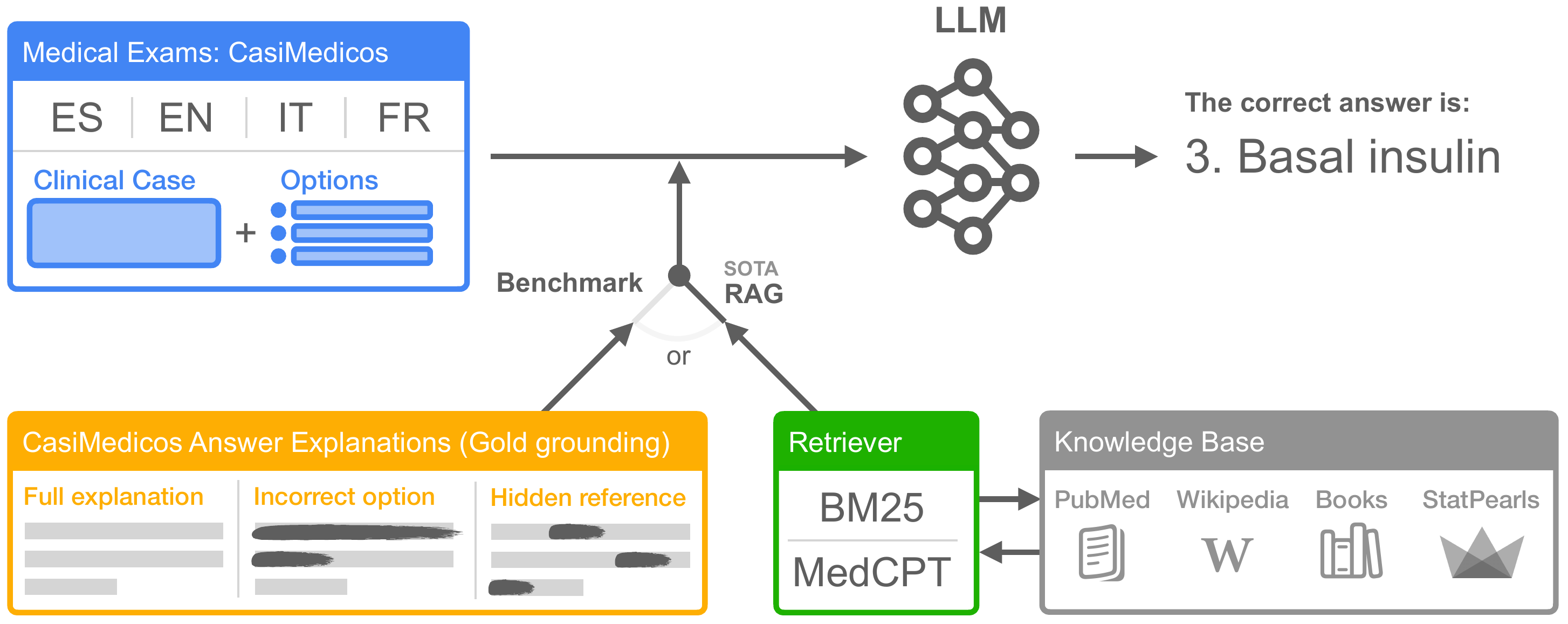} 
\caption{Graphical description of the MedExpQA benchmark in which various types of gold and external medical knowledge are added to Large Language Models in order to find the correct answer in the CasiMedicos dataset.}
\label{fig:benchmark_overview}
\end{figure}

In this paper we present MedExpQA (Medical Explanation-based Question Answering), which is, to the best of our knowledge, the first multilingual benchmark for Medical QA. Furthermore, and unlike previous work, our new benchmark also includes gold reference explanations to justify why the correct answer is correct and also to explain why the rest of the options are incorrect. Written by medical doctors, these high-quality explanations help to assess the model's decisions based on complex medical reasoning. Moreover, our MedExpQA benchmark leverages the reference explanations as \emph{gold knowledge} to establish various upperbounds for comparison with results obtained when applying automatic MedRAG methods. By doing so, we aim to address all four shortcomings of LLMs for Medical QA listed above.

Although by design independent of the specific source data used, for this work we leverage the Antidote CasiMedicos dataset \citep{Agerri2023HiTZAntidoteAE,goenaga2023explanatory}, which consist of Resident Medical Exams or \emph{M\'edico Interno Residente} in Spanish, an exam similar to other licensing examinations such as USMLE, to setup MedExpQA. In addition to a short clinical case, a question and the multiple-choice options, CasiMedicos includes gold reference explanations regarding both the correct and incorrect options. Originally in Spanish, CasiMedicos was translated and annotated in English, French and Italian \citep{goenaga2023explanatory}.

Figure \ref{fig:benchmark_overview} provides an overview of the MedExpQA benchmark. Taking CasiMedicos as the data source, the basic input, without any additional knowledge, to the LLM consists of a clinical case and the multiple-choice options. Furthermore, the model can also be provided with three types of gold reference explanations (or gold knowledge grounding) extracted from the CasiMedicos explanations: (i) the full gold explanation as written by the medical doctors;  (ii) only the explanations regarding the incorrect answers and, (iii) the full gold explanation with explicit references to the possible answers hidden. Finally, we can also apply automatic knowledge retrieval approaches such as \textsc{MedRAG} to provide LLMs with automatically obtained up-to-date medical knowledge. Thus, in MedExpQA it is possible to compare not only whether the \textsc{MedRAG} methods improve over the basic input with no external knowledge added, but also to establish the differences in performance of LLMs (with or without RAG) with respect to results obtained when gold reference explanations are available. An additional benefit of MedExpQA being multilingual is that we get to compare LLMs performance not only for English, but also on popular languages such as Spanish, French or Italian.

\begin{figure*}[h!]
    \centering
         \includegraphics[width=0.8\textwidth]{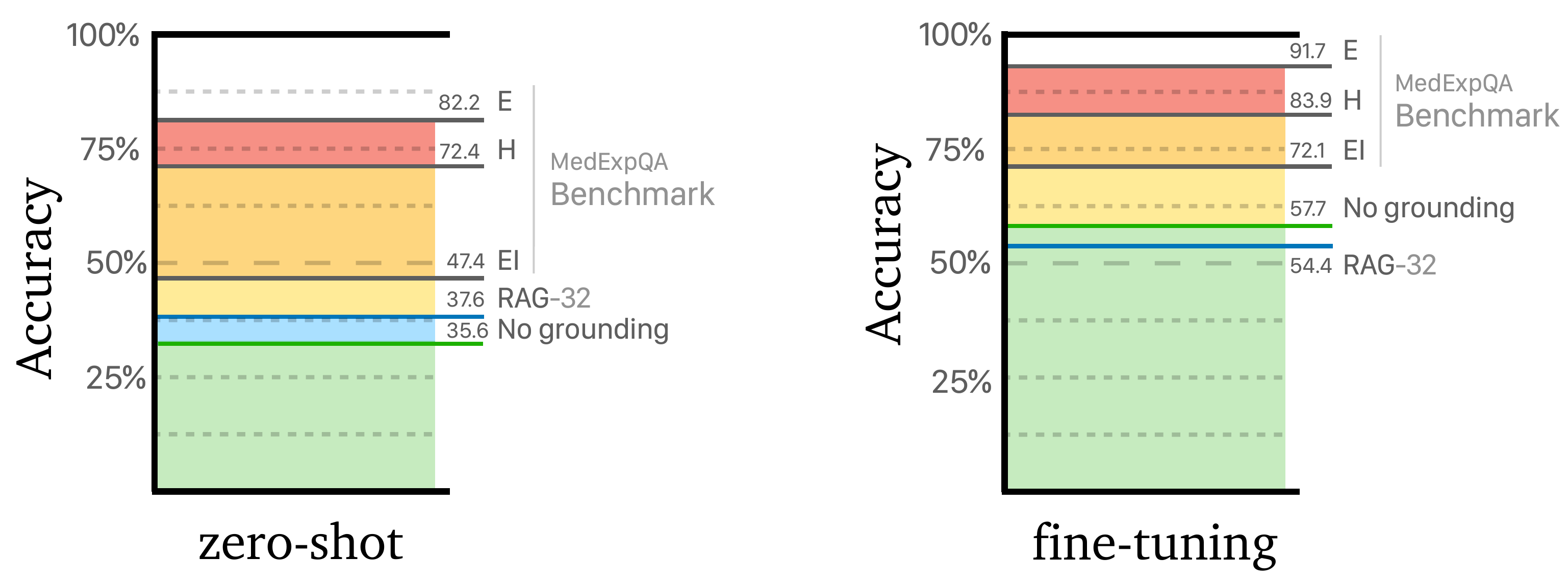}
    \caption{Overview of averaged results in MedExpQA for gold and automatically knowledge grounding based on Retrieval Augmented Generation (RAG). \textcolor{black}{\textit{E}: gold explanations written by medical doctors; \textit{H}: E with explicit references to the possible answers hidden; and \textit{EI}: gold explanations about the incorrect options; \textit{RAG-32}: automatically retrieved knowledge grounding (details in Section \ref{sec:experimental}); \textit{no-grounding}: baseline model with no external knowledge.}}
    \label{fig:benchmark}
\end{figure*}

Figure \ref{fig:benchmark} shows that comprehensive multilingual experimentation on MedExpQA using four state-of-the-art LLMs including LLaMA \citep{touvron2023llama} PMC-LLaMA \citep{wu2023pmcllama}, Mistral \citep{jiang2023mistral} and BioMistral \citep{labrak2024biomistral}, demonstrate that LLMs performance, even when improved with external knowledge from \textsc{MedRAG} \textcolor{black}{(corresponding to RAG-32 in Figure \ref{fig:benchmark})}, still has a long way to go to get closer to the performance obtained when the external knowledge available to the LLM is based on gold reference explanations \textcolor{black}{(\textit{E} and \textit{H} in Figure \ref{fig:benchmark})}. Another interesting point is that fine-tuning results in huge performance increases across settings and models but at the cost of making \textsc{MedRAG} redundant. In other words, \textsc{MedRAG} only has a positive impact in zero-shot settings. We believe that this illustrates the difficulty of automatically retrieving and integrating readily available knowledge in a way that may positively impact final downstream results on Medical QA. Finally, results are substantially lower for French, Italian and Spanish, which suggests that more work is needed to improve LLMs performance for languages different to English. Summarizing, the main contributions of our work are the following:

\begin{enumerate}
\item MedExpQA: the first multilingual benchmark for MedicalQA including gold reference explanations.
\item Comprehensive study on the role of medical knowledge to answer medical exams by leveraging gold reference explanations written by medical doctors as upper bound with respect to automatically retrieved knowledge using state-of-the-art RAG techniques.
\item Experimental results demonstrate that fine-tuning clearly outperforms querying the LLMs in zero-shot, making redundant the external knowledge obtained via RAG.
\item Overall performance of LLMs with or without RAG still has large room for improvement when compared with any of the results obtained using gold reference explanations.
\item Performance for French, Italian and Spanish substantially lower for every LLM in every evaluation setting, which stresses the urgent need of advancing the state-of-the-art for Medical QA in languages different to English. 
\item Data, code and fine-tuned models available to facilitate reproducibility of results and benchmarking of LLMs in the medical domain\footnote{\url{https://huggingface.co/datasets/HiTZ/MedExpQA}}.
\end{enumerate}

\textcolor{black}{In the rest of the paper we first discuss the related work and then in Section \ref{sec:Materials} we describe the Large Language Models (LLM) and the Retrieval Augmented Generation method used for experimentation. Section \ref{sec:dataset} provides a description of the MedExpQA benchmark, including the Antidote CasiMedicos dataset. The experimental setup is explained in Section \ref{sec:experimental} and results are reported in Section \ref{sec:results}. Section \ref{sec:discussion} offers a discussion of the main issues raised by the empirical results obtained. We finish with some concluding remarks and future work in Section \ref{sec:conclusions}.}

\section{Related Work}\label{sec:Related}

We are currently seeing a vertiginous rhythm in the development of Large Language Models (LLMs) which is having a great impact on
Natural Language Processing for the medical domain. This is particularly true on Medical Question Answering tasks where LLMs have been successfully applied to generate answers to highly specialized medical questions. Thus, the performance improvements on Abstractive Medical Question Answering of general purpose LLMs such GPT-4 \citep{nori2023capabilities} and GPT-3 \cite{brown2020language}, PaLM \cite{Chowdhery2022PaLMSL}, LLaMa \cite{touvron2023llama} or Mistral \citep{jiang2023mistral}, has resulted in a huge interest to adapt or to generate LLMs specialized for medical text processing.

Some of these models are based on the encoder-decoder architecture, such as SciFive \cite{scifive}, and English T5 model adapted to the scientific domain, or Medical-mT5, a multilingual model built by fine-tuning mT5 on a multilingual corpus of 3B tokens \citep{medical-mt5}. However, the large majority of the LLMs specially generated for medical applications are autorregresive decoder models such as BioGPT \cite{biogpt}, ClinicalGPT \cite{Wang2023ClinicalGPTLL}, Med-PaLM \cite{singhal2023large}, MedPaLM-2 \citep{singhal2023towards}, PMC-LLaMA \cite{wu2023pmcllama}, and more recently, BioMistral \citep{labrak2024biomistral}.

\textcolor{black}{These models have been reporting high-accuracy scores on various medical QA benchmarks, which generally consist of exams or general medical questions. Several of the most popular Medical QA datasets \citep{jin-etal-2019-pubmedqa,abacha2019overview,vilares2019head,abacha2019bridging,jin2021disease,pal2022medmcqa} have been grouped into two multi-task English benchmarks, namely, MultiMedQA \citep{singhal2023large} and \textsc{MIRAGE} \citep{xiong2024benchmarking} with the aim of providing an easier comprehensive experimental evaluation benchmark of LLMs for Medical QA.}

Despite recent improvements on these benchmarks that had led to claims about the capacity of LLMs to encode clinical knowledge \citep{singhal2023large}, these models remain hindered by well known issues related to: (i) their tendency to generate plausible-looking but factually inaccurate answers and, (ii) working with outdated knowledge as their pre-training data may not be up-to-date to the latest available medical progress; (iii) the large majority of these benchmarks do not include gold reference explanations to help evaluate the reasoning capacity of LLMs to predict the correct answers; (iv) they have mostly been developed for English, which leaves a huge gap regarding the evaluation of the abilities of LLMs for other languages.

Regarding the first issue listed above, it should be considered that these LLMs are not restricted to the input context to generate the answer as they are able to produce word by word output by using their entire vocabulary in an auto-regressive manner \citep{raffel2020exploring}. This often results in answers that are apparently plausible and factually correct, when in fact they are not always factually reliable. With respect point (ii), while LLMs are pre-trained with large amounts of texts, they may still lack the specific knowledge required to answer highly specialized questions or it may simply be in need of an update.

Recent work \citep{zakka2024almanac} has proposed Retrieval Augmented Generation (RAG) \citep{lewis2020retrieval} to mitigate these limitations. This method involves incorporating relevant external knowledge into the input of these LLMs with the aim of improving the final generation. By doing so, it increases the probability of generated responses being grounded in the automatically retrieved evidence, thereby enhancing the accuracy and quality of the output. Some of the most common retrieval methods employed include TF-IDF, BM25 \citep{bm25}, and others more specific to the medical domain such as MedCPT \citep{btad651}. With the aim of providing an exhaustive evaluation of RAG methods for the medical domain, the \textsc{MIRAGE} benchmarch includes 5 well-known English Medical QA datasets which are used to compare zero-shot performance of various LLMs whenever automatically retrieved knowledge is available via their \textsc{MedRAG} method or in the absence of it. According to the authors, \textsc{MedRAG} not only helps to address the problem of hallucinated content by grounding the generation on specific contexts, but it also provides relevant up-to-date knowledge that may not be encoded in the LLM \citep{xiong2024benchmarking}. By employing \textsc{MedRAG} they are able to clearly improve the zero-shot results of some of the LLMs tested, although for others results are rather mixed.

Finally, and to the best of our knowledge, no Medical QA benchmark currently addresses the last two shortcomings, namely, the lack of gold reference explanations and multilinguality. Motivated by this, we propose MedExpQA, a multilingual benchmark including gold reference explanations written by medical doctors that can be leveraged to setup various upperbound results to be compared with the performance of LLMs enhanced by automatic RAG methods.

\section{Materials and Methods}\label{sec:Materials}

In this section we describe the main resources used in our experimentation with MedExpQA, namely, the Large Language Models (LLMs) tested on our benchmark and \textsc{MedRAG}, the Retrieval Augmented Generation method proposed by \citet{xiong2024benchmarking} to automatically retrieve medical knowledge.

\subsection{Models}\label{subsubsec:models}

We selected two open source state-of-the-art LLMs in the MedicalQA domain at the time of writing: PMC-LLaMA \citep{wu2023pmcllama} and BioMistral \citep{labrak2024biomistral}. 

PMC-LLaMA is based on LLaMA \citep{touvron2023llama}, one of the most popular LLMs currently available. PMC-LLaMA is an open-source language model specifically designed for medical applications. This model was first pre-trained on a combination of PubMed-related English academic papers from the S2ORC corpus \citep{lo-etal-2020-s2orc} and from medical textbooks. It was then further fine-tuned on a dataset of instruction-based medical texts. For our experiments we pick the 13B parameter variant of this model which outperforms LLaMA-2 \citep{touvron2023llama}, Med-Alpaca \citep{han2023medalpaca}, and Chat-Doctor \citep{li2023chatdoctor} in various Medical QA tasks including MedQA \citep{jin2021disease}, MedMCQA \citep{pmlr-v174-pal22a}, and PubMedQA \citep{jin-etal-2019-pubmedqa}.

BioMistral \citep{labrak2024biomistral} is a suite of open-source models based on Mistral \citep{jiang2023mistral} further pre-trained using English textual data from PubMed Central Open Access \footnote{PMC Open Access Subset. Available from \url{https://www.ncbi.nlm.nih.gov/pmc/tools/openftlist/}}. They released a set of 7b parameter models following merging techniques like TIES \citep{yadav2023tiesmerging}, DARE \citep{yu2023language}, and SLERP \citep{slerp1985}. In this paper we use the DARE variant of BioMistral as it is the best performing model on the MedQA benchmark, outscoring other state-of-the-art LLMs on Medical QA evaluations, including PMC-LLaMA.

Additionally, and in order to contrast their performance against their general purpose counterparts, we also test LLaMA-2 and Mistral. Thus, for both PMC-LLaMa and LLaMA-based models we use the 13 billion parameter variants. As BioMistral is only available in the 7b version, we also pick the Mistral model of 7b parameters.

Every zero-shot and fine-tuning experiment with LLMs are performed via the HuggingFace API \citep{wolf-etal-2020-transformers}.

\subsection{Retrieval-Augmented Generation (RAG)}\label{subsubsec:RAG}

We apply \textsc{MedRAG} as the Retrieval-Augmented Generation (RAG) state-of-the-art technique especially developed for the medical domain \citep{xiong2024benchmarking}. RAG approaches are mostly composed of three components: the LLM, the retrieval method and the data source from which to retrieve the knowledge. \textsc{MedRAG} includes four retrievers and four different corpora as data sources. 

With respect the retrievers, we use both BM25 \citep{bm25} and MedCPT \citep{btad651} to perform the retrieval and fuse the retrieved candidate lists into one using Reciprocal Rank Fusion (RRF) \citep{rrf}. BM25 is a ranking function used in Information Retrieval to rank documents based on their relevance to a given query. It combines Term Frequency (TF) and Inverse Document Frequency (IDF) to calculate the relevance score of a document to a query taking into account the document length for normalization. MedCPT is a Contrastive Pre-trained Transformer model trained with PubMed search logs for zero-shot biomedical information retrieval. This model retrieves the relevant documents in the knowledge base considering relationships between different medical entities and concepts in the query.

Regarding the data sources, we use \textsc{MedCorp}, a combination of the four corpora available in \textsc{MedRAG}: PubMed, Textbooks \citep{jin2021disease} for domain-specific knowledge, StatPearls\footnote{\url{https://www.statpearls.com/}} for clinical decision support, and Wikipedia for general knowledge. According to the \textsc{MIRAGE} results \citep{xiong2024benchmarking}, using \textsc{MedCorp} was the only realistic option for \textsc{MedRAG} to systematically improve results over the baseline for most of the LLMs and retriever methods evaluated.

\section{MedExpQA: A new multilingual benchmark for Medical QA}\label{sec:dataset}

Although independently designed with respect to any specific dataset, in this paper we setup MedExpQA, introduced in Section \ref{sec:benchmark}, on the Antidote CasiMedicos dataset \citep{Agerri2023HiTZAntidoteAE,goenaga2023explanatory}, which is described in detailed in Section \ref{sec:casimedicos}.

\begin{table}[!h]
\centering
\scriptsize
\begin{tabular}{l |p{10.5cm}}
\toprule
\midrule
\textbf{C} & 30-year-old man with no past history of interest. He comes for consultation due to the presence of small erythematous-violaceous lesions that on palpation appear to be raised in the pretibial region. The analytical study shows a complete blood count and coagulation study without alterations, and in the biochemistry, creatinine and ions are also within the normal range. The urinary sediment study shows hematuria, for which the patient had already been studied on other occasions, without obtaining a definitive diagnosis. Regarding the entity you suspect in this case, it is FALSE that \\ \midrule
\textbf{O} & \begin{tabular}[c]{@{}p{10.5cm}}\textbf{(1)} In 20 to 50\% of cases there is elevation of serum IgA concentration.\\ \textbf{(2)} In the renal biopsy the mesangial deposits of IgA are characteristic. \\ \textbf{(3)} It is frequent the existence of proteinuria in nephrotic range.\\ \textbf{(4)} It is considered a benign entity since less than 1/3 of patients progress to renal failure.\\\textbf{(5)} The cutaneous biopsy allows to establish the diagnosis in up to half of the cases. \end{tabular}  \\
\midrule
\textbf{A}  & \textbf{3}\\ 
\midrule
\midrule
\textbf{E} & They are talking to us with high probability of a mesangial IgA glomerulonephritis or Berger's disease. Therefore, we are going to discard options one by one: \textcolor{black}{1: True. Serum IgA elevation is found in up to 50\% of cases.} \textcolor{black}{2: True. Mesangial IgA deposits are present in almost 100\% of cases.} \textcolor{black}{3: This option is false, because this glomerulonephritis is classically manifested with nephritic and not nephrotic syndrome (although in some rare cases proteinuria in nephrotic range does appear, but in the MIR they do not ask about these rare cases).} \textcolor{black}{4: At the beginning this option generated doubts in me, but looking in the literature, it is true that the evolution to renal failure (according to last series) occurs in about 25\% of the cases, so this option is true.} \textcolor{black}{5: Skin biopsy, because it is easier to perform than renal biopsy, is the diagnostic technique of choice (the skin lesions that constitute Schonlein-Henoch purpura, so frequently associated with this entity and which the patient in the case presents, are biopsied).}\\ 
\midrule

\textbf{EC} & \textcolor{black}{3: This option is false, because this glomerulonephritis is classically manifested with nephritic and not nephrotic syndrome (although in some rare cases proteinuria in nephrotic range does appear, but in the MIR they do not ask about these rare cases).}\\
\midrule
\textbf{EI} & \textcolor{black}{1: True. Serum IgA elevation is found in up to 50\% of cases.} \textcolor{black}{2: True. Mesangial IgA deposits are present in almost 100\% of cases.}  \textcolor{black}{4: At the beginning this option generated doubts in me, but looking in the literature, it is true that the evolution to renal failure (according to last series) occurs in about 25\% of the cases, so this option is true.} \textcolor{black}{5: Skin biopsy, because it is easier to perform than renal biopsy, is the diagnostic technique of choice (the skin lesions that constitute Schonlein-Henoch purpura, so frequently associated with this entity and which the patient in the case presents, are biopsied).}\\
\bottomrule
\end{tabular}
\caption{Document in the Antidote CasiMedicos dataset with the correct and incorrect explanations manually annotated. \textbf{C}: Clinical case and question; \textbf{O}: Multiple-choice options; \textbf{A}: Correct answer; \textbf{E}: Full gold reference explanation written by medical doctors; \textbf{EC}: Explanation about the correct answer; \textbf{EI}: Explanation about the incorrect answers.}
\label{tab:documentexample}
\end{table}

\subsection{Antidote CasiMedicos Dataset}\label{sec:casimedicos}

Every year the Spanish Ministry of Health releases the previous year's Resident Medical exams or \emph{M\'edico Interno Residente} (MIR) which, as depicted in Table \ref{tab:documentexample}, include a clinical case (\textbf{C}), the multiple choice options (\textbf{O}), and the correct answer (\textbf{A}). The MIR exams are then commented every year by the CasiMedicos MIR Project 2.0\footnote{\url{https://www.casimedicos.com/mir-2-0/}} which means that CasiMedicos medical doctors voluntarily write gold reference explanations (full gold explanation \textbf{E} in Table \ref{tab:documentexample}) providing reasons for both correct (\textbf{EC}) and incorrect options (\textbf{EI}).

The Antidote CasiMedicos dataset \citep{Agerri2023HiTZAntidoteAE,goenaga2023explanatory} consists of the original Spanish commented exams which were cleaned, structured and manually annotated to link the relevant textual parts in the gold reference explanation (\textbf{E}) with the correct (\textbf{EC}) or incorrect options (\textbf{EI}). Once the Spanish version of the dataset was created,  parallel translated annotated versions were generated for English, French, and Italian.

A quantitative description of the multilingual Antidote CasiMedicos dataset is given in Table \ref{tab:MultilingQuantitative}. 
The average number of tokens in the clinical cases is 137, being quite similar for Spanish and Italian (140.3 and 142.2 respectively), while for English the average is smaller (115.4 tokens) while the French one is the largest (150.1 tokens). The average length in tokens of the multiple choice options (79.6 tokens in average) is quite high but with a high variability. The multiple choice options may consist of short drug names (the minimum number of words is around 15-17) to long descriptions of treatments or medical claims as illustrated by the example shown in Table \ref{tab:documentexampleComplex}. The full gold reference explanations that professional medical doctors write can be quite long (170.25 tokens in average) but it should be noted that some documents lack the explanation about the correct answer.

\begin{table*}[!h]
\centering
\small
\begin{tabular}{l|l|r|r|r}
\toprule
&\multicolumn{1}{c|}{\textbf{Number of tokens}} & \multicolumn{1}{c|}{\textbf{Average}} & \multicolumn{1}{c|}{\textbf{Min}} & \multicolumn{1}{c}{\textbf{Max}} \\
\midrule
\multirow{5}{*}{\textbf{Spanish}} &
Clinical Case (C) & 140.3\textcolor{GrayD}{~$\pm$~ 62.4} & 41 & 504 \\
&Multiple choice options (O) & 77.0 \textcolor{GrayD}{~$\pm$~47.0} & 15 & 297 \\
&Explanation about the correct (EC) & 58.9 \textcolor{GrayD}{~$\pm$~37.7} & 0 & 483\\
& Full explanation (E) & 174.1\textcolor{GrayD}{~$\pm$~147.8} & 9 & 982 \\
\midrule
\multirow{5}{*}{\textbf{English}} &
Clinical Case (C)  & 115.4\textcolor{GrayD}{~$\pm$~52.8}& 34 & 419 \\
&Multiple choice options (O) & 64.7\textcolor{GrayD}{~$\pm$~37.1} & 15 & 217  \\
&Explanation about the correct (EC) & 47.3\textcolor{GrayD}{~$\pm$~30.4} &  0 & 382 \\
&Full explanation (E) & 139.1\textcolor{GrayD}{~$\pm$~117.7} & 4 & 784\\
\midrule
\multirow{5}{*}{\textbf{Italian}} &
Clinical Case (C) & 142.2\textcolor{GrayD}{~$\pm$~64.5} & 35 & 539 \\
&Multiple choice options (O) & 79.0\textcolor{GrayD}{~$\pm$~50.1} & 17 &  284 \\
&Explanation about the correct  (EC) & 60.6\textcolor{GrayD}{~$\pm$~38.4} & 0 &  500\\
&Full explanation (E) & 179.1\textcolor{GrayD}{~$\pm$~150.6} & 8 & 1013\\
\midrule
\multirow{5}{*}{\textbf{French}} &
Clinical Case (C) & 150.1\textcolor{GrayD}{~$\pm$~68.6} & 39 & 586 \\
&Multiple choice options (O)& 83.0\textcolor{GrayD}{~$\pm$~52.8} & 16 & 319  \\
&Explanation about the correct  (EC) & 63.9\textcolor{GrayD}{~$\pm$~41.2} & 0 & 535\\
&Full explanation (E) & 188.7\textcolor{GrayD}{~$\pm$~158.9} & 8 & 1076\\
\midrule
\multirow{5}{*}{\textbf{Avg. ALL}} &
Clinical Case (C) & 137 &  &  \\
&Multiple choice options (O)& 79.6 & &   \\
&Explanation about the correct  (EC) & 57.6& &  \\
&Full explanation (E)  & 170.25 & & \\
\bottomrule
\end{tabular}
\caption{Quantitative description of the multilingual CasiMedicos dataset. Number of tokens in the clinical case including: the question (C), the multiple-choice options (O), the explanation about the correct answer (EC) and the full gold reference explanation (E) including argumentation about the correct and incorrect answers.}
\label{tab:MultilingQuantitative}
\end{table*}

The complexity of some of the clinical case questions can be appreciated in the example shown in Table \ref{tab:documentexampleComplex} where the possible answers (section \textbf{O}) describe disorders (option (1)), treatments (options (2) and (3)) or medical statements (options (4) and (5)). Furthermore, while in the majority of the cases the question is about the correct answer, sometimes the required option is the incorrect one, as shown in Tables \ref{tab:documentexample} and \ref{tab:documentexampleComplex}.

\begin{table}[!h]
\centering
\scriptsize
\begin{tabular}{l |p{10.5cm}}
\toprule
\multicolumn{1}{c}{} & \multicolumn{1}{c}{\textbf{Example of a document from the CasiMedicos Dataset}} \\ \midrule 
\textbf{C} & A 63-year-old woman comes to the emergency department reporting severe headache with signs of meningeal irritation, bilateral visual disturbances and ophthalmoplegia. A CT scan showed a 2 cm space-occupying lesion in the sella turcica compatible with pituitary adenoma with signs of intratumoral hemorrhage, with deviation of the pituitary stalk and compression of the glandular tissue.  Mark which of the following answers is WRONG: \\ \midrule
\textbf{O} & \begin{tabular}[c]{@{}p{10.5cm}}\textbf{(1)} Diagnostic suspicion is pituitary apoplexy.\\ \textbf{(2)} Treatment with high-dose corticosteroids should be initiated and the evolution observed, since this treatment could reduce the volume of the lesion and avoid intervention. \\ \textbf{(3)} Treatment with glucocorticoids should be considered to avoid secondary adrenal insufficiency that would compromise the patient's vital prognosis.\\ \textbf{(4)} The presence of ophthalmoplegia and visual defects are indications for prompt intervention by urgent surgical decompression. \\\textbf{(5)} After resolution of the acute picture, the development of panhypopituitarism is frequent.\end{tabular}  \\
\midrule
\textbf{A}  & \textbf{4}\\ 
\bottomrule
\end{tabular}
\caption{Example of a document in the CasiMedicos dataset with very different types of response options. (1) diagnosis; (2) and (3) treatments; and (4) and (5) correspond to medical statements.}
\label{tab:documentexampleComplex}
\end{table}

The final Antidote CasiMedicos Dataset consists of 622 documents per language \citep{Agerri2023HiTZAntidoteAE,goenaga2023explanatory}. The dataset official distribution already provide train, validation and test splits\footnote{\url{https://huggingface.co/datasets/HiTZ/casimedicos-exp}} (depicted in Table \ref{tab:DatasetDistribution}), which we use for the all the experiments presented in Section \ref{sec:results}.

\begin{table*}[!h]
\centering
\begin{tabular}{l|ccc|}
\cline{2-4}
& Train & Validation & Test \\
\hline
\multicolumn{1}{|l|}{Clinical cases} 	& 434 & 63 & 125 \\
\hline
\multicolumn{1}{|l|}{Total} & \multicolumn{3}{c|}{622} \\ 
\hline
\end{tabular}
\caption{Number of documents in CasiMedicos train, validation and test splits.}\label{tab:DatasetDistribution}
\end{table*}

\begin{figure*}[!htbp]
    \centering
    \begin{subfigure}[t]{0.3\textwidth}
        \centering
         \includegraphics[width=\textwidth]{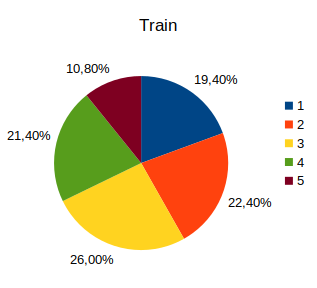}
         \caption{\centering Distribution of correct answers in the \textbf{train} dataset.}\label{fig:CorrectTrain}
    \end{subfigure}%
    \quad
    \begin{subfigure}[t]{0.3\textwidth}
        \centering
         \includegraphics[width=\textwidth]{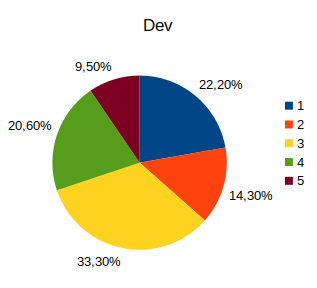}
         \caption{\centering Distribution of correct answers in the \textbf{validation} dataset.}\label{fig:CorrectDev}
    \end{subfigure}
    \quad
    \begin{subfigure}[t]{0.3\textwidth}
        \centering
         \includegraphics[width=\textwidth]{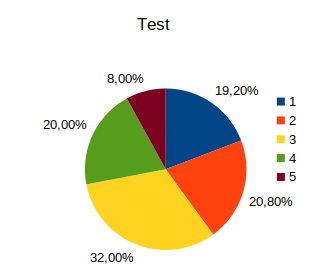}
         \caption{\centering Distribution of correct answers in the \textbf{test} dataset.}\label{fig:Test}
    \end{subfigure}
    \caption{Distribution of correct answers in the train, validation and test splits. The percentage in blue indicates the proportion of exams with the first option, number 1, as correct answer; orange corresponds to option 2; yellow to option 3; green to option 4; and brown to option 5. Note that not every document includes 5 possible options.}
    \label{fig:CorrectAnswerDistribution}
\end{figure*}

Finally, we examined the distribution of correct answers in each of the three splits (train, validation and test) to consider the possibility that an unbalanced distribution might condition the results of the tested models. Figure \ref{fig:CorrectAnswerDistribution} shows that, although most of the exams have the option 3 as the correct answer, the distribution among the correct answers in the three subsets is quite balanced. This suggests that this particular issue should not influence the final experimental results.

\subsection{The MedExpQA Benchmark}\label{sec:benchmark}

MexExpQA is a multilingual benchmark to evaluate LLMs in Medical Question Answering. Unlike previous work, MedExpQA includes reference gold explanations written by medical doctors which are leveraged to setup a benchmark with three types of gold knowledge: (i) the full gold reference explanation (part \textbf{E} in Table \ref{tab:documentexample}); (ii) the full gold reference explanation corresponding to the incorrect options only (\textbf{EI}) and (iii), the full gold reference explanation masking the explicit references in the text to the multiple-choice options.

In other words, and as illustrated in Figure \ref{fig:benchmark_overview}, we use these three types of high-quality explanations written by medical doctors as a proxy of relevant gold knowledge that may be used by LLMs to answer medical questions. Thus, the results obtained by LLMs with each type of gold knowledge can be seen as the upperbound results provided by our benchmark to establish how well LLMs can perform according to the different types of specialized gold knowledge readily available. In the following we describe in detail each of the three types of gold reference explanations that we generate to setup our benchmark.



\subsubsection{Full Reference Gold Explanations}\label{sec:gold-upperbounds}

The full explanation (\textbf{E}) about the correct and incorrect answers is given as context to the LLM, in what we assume to be gold specific knowledge for the model to answer the medical questions of CasiMedicos. Being the full gold reference explanation, we consider this to be the best possible form of gold knowledge that we can provide the LLM with. In other words, the performance obtained in MedExpQA using this type of knowledge will mark the upperbound for this particular benchmark. Table \ref{tab:full} provides an example of the full gold reference explanation for the same document already discussed in Table \ref{tab:documentexample}.

\begin{table}[!h]
\centering
\scriptsize
\begin{tabular}{l |p{10.5cm}}
\toprule
\textbf{E} & They are talking to us with high probability of a mesangial IgA glomerulonephritis or Berger's disease. Therefore, we are going to discard options one by one: \textcolor{black}{1: True. Serum IgA elevation is found in up to 50\% of cases.} \textcolor{black}{2: True. Mesangial IgA deposits are present in almost 100\% of cases.} \textcolor{black}{3: This option is false, because this glomerulonephritis is classically manifested with nephritic and not nephrotic syndrome (although in some rare cases proteinuria in nephrotic range does appear, but in the MIR they do not ask about these rare cases).} \textcolor{black}{4: At the beginning this option generated doubts in me, but looking in the literature, it is true that the evolution to renal failure (according to last series) occurs in about 25\% of the cases, so this option is true.} \textcolor{black}{5: Skin biopsy, because it is easier to perform than renal biopsy, is the diagnostic technique of choice (the skin lesions that constitute Schonlein-Henoch purpura, so frequently associated with this entity and which the patient in the case presents, are biopsied).}\\ 
\bottomrule
\end{tabular}
\caption{Full explanation (E) of the example in Table \ref{tab:documentexample}. The explanation about the correct answer is marked in blue and the remaining 4 explanations for the incorrect options in green. }
\label{tab:full}
\end{table}

\subsubsection{Explanation of the Incorrect Options}

As shown in Table \ref{tab:EI}, for this particular type of gold knowledge we only use the part of the full gold reference explanation corresponding to the explanations about the incorrect options (\textbf{EI}). This type gold knowledge aims to test the capacity of LLMs to correctly answer the medical question by knowing which options are incorrect.

\begin{table}[!h]
\centering
\scriptsize
\begin{tabular}{l |p{10.5cm}}
\toprule
\textbf{EI} & They are talking to us with high probability of a mesangial IgA glomerulonephritis or Berger's disease. Therefore, we are going to discard options one by one: \textcolor{black}{1: True. Serum IgA elevation is found in up to 50\% of cases.} \textcolor{black}{2: True. Mesangial IgA deposits are present in almost 100\% of cases.}  \textcolor{black}{5: Skin biopsy, because it is easier to perform than renal biopsy, is the diagnostic technique of choice (the skin lesions that constitute Schonlein-Henoch purpura, so frequently associated with this entity and which the patient in the case presents, are biopsied).}\\ 
\bottomrule
\end{tabular}
\caption{Explanation of the Incorrect Options (EI) which corresponds to the full explanation (E) of the example in Table \ref{tab:documentexample} with the explanation of the correct answer removed.}
\label{tab:EI}
\end{table}



Depending on the nature of the question, sometimes medical doctors consider sufficient to only explain the correct answer. Thus, it should be noted that not every document in CasiMedicos includes the gold reference explanations about the incorrect options. On average, 20.5\% of the explanations correspond in their entirety to the correct answer (17.7\% in the train set, and 22.2\% and 21.6\% in the validation and test, respectively), \textcolor{black}{while 26.7 include the explanations for all the possible options}.
Obviously, as CasiMedicos is a multilingual parallel dataset, this phenomenon occurs across the four languages: English, French, Italian and Spanish.

\subsubsection{Full Gold Explanation with Explicit References Hidden}

As it can be appreciated in the full gold reference explanations discussed above, most of the time medical doctors provide explicit textual references regarding the correct or incorrect options. In order to analyze the impact of these explicit signals or patterns on the LLMs performance, we decided to mask those explicit references to establish how well LLMs could answer with actual gold knowledge but without the easy clues in the text pointing to the correct or incorrect answers.

In order to avoid the manual annotation of 2488 documents, we prompt GPT-4\footnote{\texttt{gpt-4-1106-preview}} \citep{openai2024gpt4} with a set of rules and in-context-learning examples to automatically mask the specific areas of text that may point the model at the correct or incorrect answer without any further reasoning. The prompt can be found in \ref{sec:prompts}, Figure \ref{fig:pixt3_overview}.

\textcolor{black}{A small manual analysis of a subset of GPT-4-generated texts revealed a strong correlation with human annotations. To further validate the efficacy of our method, we randomly selected 80 documents (20 per language) and measured performance across the four languages. This resulted in an average F1 score of 0.85 with a standard deviation of 0.02.}

Thus, this method allowed us to perform this rather precise multilingual redacting process over the 2488 documents in a fast and cost effective manner. Table \ref{tab:hidden} shows how every explicit reference to the correct or incorrect answers discussed previously now appear as \textbf{[\textsc{hidden}]}.
 
 \begin{table}[!h]
\centering
\scriptsize
\begin{tabular}{l |p{10.5cm}}
\toprule
\textbf{H} & They are talking to us with high probability of a mesangial IgA glomerulonephritis or Berger's disease. Therefore, we are going to discard options one by one: \textcolor{black}{1: True. Serum IgA elevation is found in up to 50\% of cases.} \textcolor{black}{2: True. Mesangial IgA deposits are present in almost 100\% of cases.} \textcolor{black}{3: \textbf{[HIDDEN]}, because this glomerulonephritis is classically manifested with nephritic and not nephrotic syndrome (although in some rare cases proteinuria in nephrotic range does appear, but in the MIR they do not ask about these rare cases).} \textcolor{black}{4: At the beginning this option generated doubts in me, but looking in the literature, it is true that the evolution to renal failure (according to last series) occurs in about 25\% of the cases, \textbf{[HIDDEN]}.} \textcolor{black}{5: Skin biopsy, because it is easier to perform than renal biopsy, is the \textbf{[HIDDEN]} (the skin lesions that constitute Schonlein-Henoch purpura, so frequently associated with this entity and which the patient in the case presents, are biopsied).}\\ 
\bottomrule
\end{tabular}
\caption{Full gold reference explanation with explicit references hidden (\textbf{H}). Process performed by GPT-4 with the prompt in \ref{sec:prompts} Figure \ref{fig:pixt3_overview}. In this example the segments \emph{`This option is false'}, \emph{`so this option is true'} and \emph{`is the diagnostic technique of choice'} are hidden.}
\label{tab:hidden}
\end{table}

The results obtained by LLMs in MedExpQA using the three types of gold knowledge described above can then be compared with other automatic knowledge retrieval approaches based, for example, on Retrieval-Augmented Generation techniques for the medical domain such as \textsc{MedRAG}, introduced in the previous section. Furthermore, we should stress that MedExpQA as a benchmark is independent of any dataset, as the only requirement is for it to include gold reference explanations of the possible answers.

\section{Experimental Setup}\label{sec:experimental}

For our experiments we selected top performing state-of-the-art models for Medical Question Answering described in Section \ref{subsubsec:models}, namely, PMC-LLaMA, LLaMA-2, BioMistral, and Mistral. 

We test these models in both zero-shot and fine-tuned settings to contrast their out-of-the-box performance against a more adjusted performance to our dataset. The models were fine-tuned using Low-Rank Adaptation (LoRA) \citep{hu2022lora}, \textcolor{black}{using adapters with a rank of 8 and a scaling factor (alpha) of 16 across all models (details about parameters used with LoRA are provided in  \ref{sec:efficiency}).} 

\textcolor{black}{The choice of hyperparameters was based on previous work using the same LLMs we use in this papers. Moreover, satisfactory results were confirmed in a preliminary round of experiments. Although these models would benefit from an exhaustive grid search of hyperparameters tailored to each model and evaluation setting, the compute required to do so exceeds the capacity of our lab.} Full details of hyperparameter settings are available in \ref{sec:hyperparameters}. \textcolor{black}{Each model was fine-tuned for 10 epochs, with checkpoints saved at the end of each. Experiments were undertaken in a NVIDIA A100 GPU (\ref{sec:efficiency} offers information about computation times). At the end of the fine-tuning process, the checkpoint with the highest performance was selected. All models underwent monolingual training using the dataset corresponding to each specific language.} We will measure the impact on MedExpQA of the different types of knowledge that LLMs may use:
\textcolor{black}{
\begin{enumerate}
\item[(i)] Gold grounding knowledge:
\begin{enumerate}
\item \textbf{E}: Full gold reference explanations as written by the medical doctors.
\item \textbf{EI}: Gold explanations about the Incorrect Options.
\item \textbf{H}: Full gold explanations with [\textsc{hidden}] explicit references to the multiple-choice options.
\end{enumerate}
\item[(ii)] Automatically obtained grounding knowledge:
\begin{enumerate}
\item \textbf{None}: Answering the medical question with no additional external knowledge.
\item \textbf{RAG-7}: Automatically obtained knowledge by applying \textsc{MedRAG} to retrieve the k=7 most relevant documents.
\item \textbf{RAG-32}: Automatically obtained knowledge by applying \textsc{MedRAG} to retrieve the k=32 most relevant documents.
\end{enumerate}
\end{enumerate}
}

We use the entire clinical case, question, and multiple-choice options to generate the query for all 6 different evaluation settings. Gold knowledge grounding is leveraged as explained in the previous section. With respect to the methods to automatically obtained external knowledge, we take into account the results obtained in the \textsc{mirage} benchmark \citep{xiong2024benchmarking} and apply M\textsc{ed}RAG by using the RRF-2 of two retrieval algorithms, namely, BM25 and MedCPT, over the \textsc{MedCorp} corpus. We use the entire clinical case, question, and multiple-choice options to generate the query to retrieve the k=7 most relevant documents. We define k=7 by computing the average token length of MedCorp documents; if we consider that 85\% of our prompts can be represented under 400 tokens, this leaves 1648 tokens for knowledge grounding, which amounts to 7 documents on average. This configuration is used to define \textbf{RAG-7}.

\begin{figure*}[h!]
    \centering
         \includegraphics[width=0.5\textwidth]{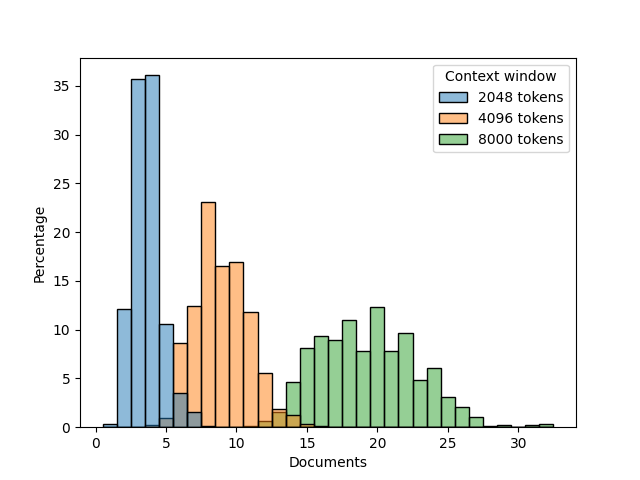}
    \caption{\textcolor{black}{Distribution of retrieved documents across different context windows. Three different histograms are shown that depict the maximum number of documents that can be accommodated within various context windows across dataset examples: 2,048 tokens (PMC-LLaMA), 4,096 tokens (LLaMA2), and 8,192 tokens (Mistral and BioMistral).}}
    
    \label{fig:doc_dist}
\end{figure*}

Furthermore, as \textsc{MedRAG} obtained best results for most of the benchmarks when retrieving at most 32 documents, we also experimented with this setting. Nevertheless, it should be considered that the context window of each model, namely, the maximum amount of word tokens that each LLM can pay attention to in the input, will determine how many of these documents are actually fed into the LLM at each forward pass. Hence, when the combination of both the retrieved documents and the prompt exceed the context window, then we truncate the amount of documents to ensure that the prompt is not affected. 
\textcolor{black}{Figure \ref{fig:doc_dist} illustrates the distribution of documents corresponding to different context window sizes. Specifically, it shows the number of examples in the dataset that align with varying numbers of retrieved documents for context windows of 2048, 4096, and 8000 tokens.} In the results reported in the next section, \textbf{RAG-32} for both zero-shot and fine-tune settings helps us to evaluate the impact of retrieving more or less relevant documents as external knowledge.  

\subsection{Evaluation}\label{subsec:evaluation}

 We ask LLMs to generate not only the index number of the predicted correct option but also the full textual answer. However, accuracy is calculated by comparing the first generated character after the prompt following \emph{``The correct answer is: ''}\footnote{And equivalent prompts for French, Italian and Spanish.}. We verify that this character always corresponds to one of the options in the exams' possible answers. \ref{sec:prompts} provides an example of the prompts used for each language and for every model. 

\section{Results}\label{sec:results}

We report the main results of the experiments performed in the MedExpQA benchmark in Table \ref{tab:BenchmarkResults1} for zero-shot while the fine-tuning accuracy scores are presented in Table \ref{tab:BenchmarkResults2}.

\textcolor{black}{
\paragraph{Zero-shot results} They show that Mistral consistently achieves the highest accuracy across every evaluation setting and language, even outscoring the medical specific BioMistral. Among the gold knowledge results, we can see that removing the explanation of the correct answer (\textbf{EI}) really hinders performance. However, using the full gold reference answer helps LLMs to obtain excellent marks. Moreover, differences between using \textbf{E} and \textbf{H} are quite large, especially for languages different to English.
}
\begin{figure}[t]
\centering
\includegraphics[scale=.3]{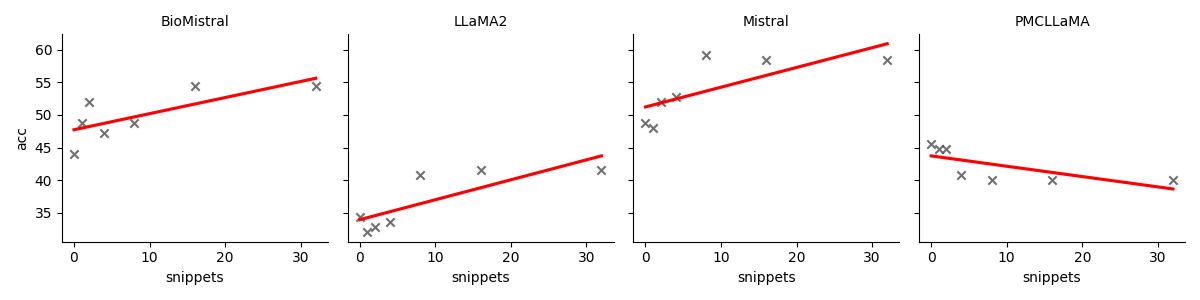} 
\caption{Performance of different models in a zero-shot setting with up to 0, 2, 4, 8, 16, and 32 retrieved snippets.}
\label{fig:scale_docs}
\end{figure}
\textcolor{black}{
It should be noted that the best automatic method still fares very badly with respect to any of the gold knowledge results, which shows that retrieval methods for the medical domain still have large room for improvement. While the best automatic method corresponds to \textbf{RAG-7}, differences in performance are not that great with respect to \textbf{None} or \textbf{RAG-32}.}

\textcolor{black}{We hypothesize that the lack of substantial improvement when using 32 snippets for knowledge grounding may indicate that a saturation point may be reached beyond which additional snippets do not provide any additional benefit. To analyze this more precisely, we conducted an evaluation of the zero-shot performance of the 4 LLMs when feeding the model from 0 to up to 32 snippets, following a power of two sequence of snippets. Thus, Figure \ref{fig:scale_docs} illustrates that a positive trend exists when increasing the number of snippets. However, we can see how this improvement tanks at around 8 snippets in most of the models. This result correlates to our findings in Tables \ref{tab:BenchmarkResults1} and \ref{tab:BenchmarkResults2}.
}
\begin{table*}[!h]
\centering
\scriptsize
\begin{tabular}{l|cccc|cccc|cccc|cccc|c}
\toprule
 & \multicolumn{4}{c}{\textbf{PMC-LLaMA}} & \multicolumn{4}{|c}{\textbf{LLaMA2}} & \multicolumn{4}{|c}{\textbf{Mistral}} & \multicolumn{4}{|c|}{\textbf{BioMistral}} & \textbf{Avg.}\\
 & \multicolumn{4}{c}{(13B)} & \multicolumn{4}{|c}{ (13B)} & \multicolumn{4}{|c}{(7B)} & \multicolumn{4}{|c|}{(7B)} & \\
\midrule
 & \textbf{EN} & \textbf{ES} & \textbf{IT} & \textbf{FR} & \textbf{EN} & \textbf{ES} & \textbf{IT} & \textbf{FR}& \textbf{EN} & \textbf{ES}& \textbf{IT} & \textbf{FR} & \textbf{EN} & \textbf{ES}& \textbf{IT} & \textbf{FR} & \textbf{ALL}\\
\midrule
\textbf{E} & 83.2 & 77.6 & 76.8 & 80.0 & 81.6 & 77.6 & 77.6 & 75.2  & \textbf{\underline{89.6}} & \textbf{\underline{88.0}} & \textbf{\underline{87.2}} & \textbf{\underline{88.0}} & 88.8 & 83.2 & 80.8 & 80.8 & \textbf{82.2}\\
\textbf{EI} & 60.0 & 42.4 & 43.2  & 46.4 & 44.0 & 31.2 & 39.2 & 44.8 & 59.2 & 53.6 & 52.0  & 52.8 & 50.4 & 44.0 & 46.4 & 49.6 & 47.4\\
\textbf{H} & 78.4 & 63.2 & 72.0 & 70.4 & 68.8 & 64.8 & 63.2 & 65.6 & 82.4 & 75.2 & 77.6 & 78.4 & 80.8  & 74.4 & 69.6 & 74.4 & 72.4\\
\midrule
\textbf{None} & 45.6 & 36.8 & 33.6 & 30.4 & 34.4 & 18.4 & 12.8 & 27.2& 48.8 & 41.6 & 40.8 & 39.2 & 44.0 & 39.2& 35.2 & 41.6 & 35.6 \\
\textbf{RAG-7} & 40.0 & 30.4 & 28.0 & 24.8 & 42.4 & 36.0* & 30.4* & 32.0 & 55.2 & \underline{44.0} & 38.4 & \underline{42.4} & 44.8 & 40.0  & 40.8 & 36.8 & \underline{37.9}\\
\textbf{RAG-32}& 40.0 & 30.4 & 28.0 & 24.8 & 41.6 & 31.2* & 32.8* & 26.4 & \underline{58.4*} & 41.6 & \underline{41.6} & \underline{42.4} & 54.4 & 37.6 & 31.2 &  39.2 & 37.6\\
\midrule
\textbf{Avg.} & 57.9 & 46.8 & 46.9 & 46.1 & 52.1 & 43.2 & 42.7 & 45.2 & \textbf{65.6} & \textbf{57.3} & \textbf{56.3} & \textbf{57.2} & 60.5 & 53.1 & 50.7 & 53.7 & -\\
\bottomrule
\end{tabular}
\caption{Zero-shot results. E: Full gold explanation. EI: Gold Explanations of the Incorrect Options; H: Full gold explanation with Hidden explicit references to the correct/incorrect answer; None: model without any additional external knowledge; RAG-7: Retrieval Augmented Generation with k=7; RAG-32: Retrieval Augmented Generation with k=32; \underline{underline}: best result per type of knowledge; \textbf{bold}: best result overall; \textbf{*}:results that are statistically significant at $\alpha = .05$ wrt to their None baseline.}
\label{tab:BenchmarkResults1}
\end{table*}

\begin{table*}[!h]
\centering
\scriptsize
\begin{tabular}{l|cccc|cccc|cccc|cccc|c}
\toprule
 & \multicolumn{4}{c}{\textbf{PMC-LLaMA}} & \multicolumn{4}{|c}{\textbf{LLaMA2}} & \multicolumn{4}{|c}{\textbf{Mistral}} & \multicolumn{4}{|c|}{\textbf{BioMistral}} & \textbf{Avg.}\\
 & \multicolumn{4}{c}{(13B)} & \multicolumn{4}{|c}{ (13B)} & \multicolumn{4}{|c}{(7B)} & \multicolumn{4}{|c|}{(7B)} & \\
\midrule
& \textbf{EN} & \textbf{ES} & \textbf{IT} & \textbf{FR} & \textbf{EN} & \textbf{ES} & \textbf{IT} & \textbf{FR}& \textbf{EN} & \textbf{ES}& \textbf{IT} & \textbf{FR} & \textbf{EN} & \textbf{ES}& \textbf{IT} & \textbf{FR} & \textbf{ALL} \\
\midrule
\textbf{E} &92.0 & 89.6 & 89.6 & 88.8 & 90.4
 & 90.4 & 89.6 & 92.0 & \textbf{\underline{94.4}} & 92.8 & 91.2 &  92.8 & \textbf{\underline{94.4}} & \textbf{\underline{93.6}} & \textbf{\underline{92.0}} &  \textbf{\underline{93.6}} & \textbf{91.7} \\
\textbf{EI} & 69.6 & 67.2 & 67.2 & 68.0 & 73.6 & 70.4 & 66.4 & 70.4 &81.6 & 78.4 & 75.2 & 76.8 & 73.6 & 72.0 & 71.2 & 71.2 & 72.1 \\

\textbf{H} & 82.4 & 76.0 &  80.0 &  82.4 & 83.2 & 85.6 & 84.0 &81.6 & 88.0 & 84.8 & 88.8 & 88.0 &  83.2 & 82.4 &  86.4 & 84.8 & 83.9\\
\midrule
\textbf{None} & 58.4
& 48.8 & 49.6 & 53.6 & 57.6 & 50.4 & 53.6 & 54.4
& 68.0 & \underline{63.2} & 56.8 & \underline{66.4} & 61.6 & 58.4& 56.8 & 65.6 & \underline{57.7} \\
\textbf{RAG-7} & 56.8 & 35.2 & 44.8 & 38.4 & 60.8 & 56.8 & 48.8 & 51.2 & 69.6 & 59.2 & 56.8 &  64.8 & 64.8 &  57.6 &  \underline{61.6} & 59.2 & 55.4\\
\textbf{RAG-32} & 56.8 & 35.2 & 44.8 &  38.4 & 60.8 & 52.0 & 51.2 & 49.6 & \underline{75.2} & 55.2 & 52.0 & 60.0& 65.6 & 57.6 & 55.2 & 60.8 & 54.4\\
\midrule
\textbf{Avg.} & 69.3 & 58.7 & 62.7 & 61.6 & 71.1 & 67.6 & 65.6 & 66.5 & \textbf{79.5} & \textbf{72.3} & 70.1 & \textbf{74.8} & 73.9 & 70.3 & \textbf{70.5} & 72.5 & -\\
\bottomrule
\end{tabular}
\caption{Fine-tuning results. E: Full gold explanation. EI: Gold Explanations of the Incorrect Options; H: Full gold explanation with Hidden explicit references to the multiple choice options; None: model without any additional external knowledge; RAG-7: Retrieval Augmented Generation with k=7; RAG-32: Retrieval Augmented Generation with k=32; \underline{underline}: best result per type of knowledge; \textbf{bold}: best result overall.}
\label{tab:BenchmarkResults2}
\end{table*}

Finally, performance on English was substantially higher for every models and RAG configurations. This manifests the English-centric focus of most LLMs while showcasing the urgent need of dedicating resources and effort to developing multilingual LLMs which could then compete across all languages included in multilingual benchmarks such as MedExpQA.

\paragraph{Fine-tuning results} They show that fine-tuning the LLMs on the CasiMedicos dataset help to greatly increase performance for every evaluation setting, language and LLM. BioMistral seems to obtain the best overall scores but that is due to its high scores on the full gold reference explanation setting (\textbf{E}). Thus, if we look at the rest of the evaluation settings, Mistral, as it happened in the zero-shot scenario, remains the best performing LLM on the MedExpQA benchmark.

The superior results of \textbf{None} with respect to RAG scores demonstrate that fine-tuning makes any external knowledge automatically retrieved using RAG methods redundant. Finally, while scores for French, Italian and Spanish remain lower than those obtained for English, performance for those languages greatly benefit from fine-tuning, especially if we compare them with their zero-shot counterpart results.

\paragraph{Overall results} Overall, results demonstrate that the gold reference explanations leveraged as knowledge for Medical QA help LLMs to obtain almost perfect scores, especially when fine-tuning the models. Fine-tuning particularly benefits \textbf{EI}, which obtains as good results as \textbf{H} applied in zero-shot settings. 

Our results allow us to draw several more conclusions. First, that despite using state-of-the-art RAG methods for the medical domain \citep{xiong2024benchmarking}, their results are rather disappointing. Both in zero-shot when compared with the results based on any kind of gold knowledge, and in fine-tuning in which RAG methods score worse than not using any additional knowledge. 

Second, our MedExpQA benchmark suggests that overall performance of even powerful LLMs such as Mistral still have a huge room for improvement to reach scores comparable to those obtained when gold knowledge is available.

\textcolor{black}{We calculated a McNemar \citep{dietterich} test of statistical significance to establish whether the \emph{RAG-7} and \emph{RAG-32} results were significantly better than their respective \emph{None} baselines. As it can be seen in Tables \ref{tab:BenchmarkResults1} and \ref{tab:BenchmarkResults2}, only five zero-shot scores (out of 64) marked with an asterisk in Table \ref{tab:BenchmarkResults1} are statistically significant at $\alpha = .05$.
}
Finally, performance for languages different to English is much lower for every model and evaluation setting. This points out to an urgent necessity to invest in the development and research of LLMs which may be optimized not only for English, but for other world languages too. Obviously, the evaluation of such LLMs would in turn require multilingual evaluation benchmarks which may be deployed to provide a comprehensive and realistic overview of their performance. We hope that contributing MedExpQA may serve as encouragement to the AI and medical research communities to generate more benchmarks of its kind for many of the world languages.

\section{Discussion}\label{sec:discussion}

The results discussed in the previous section show that even when performing fine-tuning with the full gold reference explanations LLMs  still remain several points below perfect scores. \textcolor{black}{Furthermore, the statistical analysis of the obtained results indicates that, despite differences compared to the \textbf{None} models, the performance gains (when that is the case) of models using \emph{RAG-7} or \emph{RAG-32} are, in 61 out 64 cases, not statistically significant. In contrast, the statistical analysis found out that the results using gold knowledge (\textbf{E}, \textbf{EI}, \textbf{H}) were all statistically significant at $\alpha = .05$}

Apart from the evaluation results, and in order to better understand the dataset on which the MedExpQA is setup, we performed several analysis regarding the quality and quantity of the explanations provided by the CasiMedicos medical doctors.

Regarding the quality of the explanations, we found several examples such as the one depicted in Table \ref{tab:ExplanationNotRelevant}. Instead of directly answering the question, the medical doctor (psychiatry resident) writing the explanation gives information that is not relevant to explain the correct answer (marked in red). We hypothesize that such explanations, which lack any relevant medical information, may have a negative impact on the final LLMs performance.

\begin{table}[!htbp]
\centering
\scriptsize
\begin{tabular}{l |p{10.5cm}}
\toprule
\textbf{E} & \textcolor{black}{Another simple question with an immediate answer, which offers no doubt.} It describes a patient worried about a non-existent physical defect, whose concern distresses him and prevents him from leaving the house. \textcolor{black}{As a psychiatry resident, I wish the MIR questions in my specialty were a bit more thought-provoking and in-depth, although I know that the seconds you will have saved by marking} \textbf{the fourth} \textcolor{black}{one directly are very valuable.}\\ 
\midrule
\end{tabular}
\caption{Example of a gold full explanation (E) with irrelevant and not medical comments.}
\label{tab:ExplanationNotRelevant}
\end{table}

It should be noted that, despite CasiMedicos being a high-quality dataset written voluntarily by medical doctors, sometimes (i) their explanations may not follow a repetitive formal structure and, ii) they are not always subjected to a second review by an auditor as it usually happens in specialized textual books.

Regarding the quantity of the explanations, around 5\% of the full gold reference explanations in the CasiMedicos dataset do not contain any explicit explanation regarding the correct answer. Sometimes the medical doctor explains the incorrect options, hoping that the reader may indirectly reach the correct conclusion, or sometimes they are cases such as the one discussed above.

In any case, while it is possible to filter out such examples, we thought it useful to leave them with the aim of analyzing in the future the performance of LLMs and RAG methods for these specific cases. After all, we would like LLMs to be able to also generalize in situations in which the knowledge is provided in a non-standard structured manner, as it is the case in the large majority of the full gold reference explanations provided in CasiMedicos.

We would like to give a final word on multilinguality. Results have shown that performance for French, Italian and Spanish is worse across the board and we believe that this topic has a lot of interesting questions for future research. Are these results a consequence of the pre-training of the LLMs? For the RAG experiments, how much, positive or negative, influence has the fact that the extracted knowledge from MedCorp is in English? Would it be better to prompt the model only in English and then translate the answers into each of the target languages, in what is usually known as a \emph{translate-test} approach? We believe that a benchmark such as MedExpQA would help to investigate these research questions which may be crucial to develop robust multilingual medical QA approaches.

\section{Concluding Remarks}\label{sec:conclusions}

In this paper we present MedExpQA, the first multilingual benchmark for Medical QA. As a new feature, our new benchmark also includes gold reference explanations to justify why the correct answer is correct and also to explain why the rest of the options are incorrect. The high-quality gold explanations have been written by medical doctors and they allow to test the LLMs when different types of gold knowledge is available. Comprehensive experimentation has demonstrated that automatic state-of-the-art RAG methods still have a long way to go to get near the scores obtained by LLMs when fed with gold knowledge. Furthermore, our benchmark has made explicit the lower overall performance of LLMs for languages other than English for Medical QA.

We think that MedExpQA may contribute to the development of AI tools to assist medical experts in their everyday activities by providing a robust multilingual benchmark to evaluate LLMs in Medical QA. Future work may involve evaluating LLMs not only regarding their accuracy in predicting the correct answer, but also on the quality of the explanations generated to justify such prediction. Of course, these approaches may pose new evaluation challenges that have not been yet contemplated in this work.

\section*{Acknowledgements}
We thank the CasiMedicos Proyecto MIR 2.0 for their permission to share their
data for research purposes. This work has been partially supported by the HiTZ
Center and the Basque Government (Research group funding IT1570-22). We are also thankful to several MCIN/AEI/10.13039/501100011033 projects: (i) Antidote (PCI2020-120717-2), and by European Union NextGenerationEU/PRTR; (ii) DeepKnowledge (PID2021-127777OB-C21) and ERDF A way
of making Europe; (iii)  Lotu (TED2021-130398B-C22) and European Union NextGenerationEU/PRTR; (iv) EDHIA (PID2022-136522OB-C22); (v) DeepMinor (CNS2023-144375) and European Union NextGenerationEU/PRTR. We also thank the European High Performance Computing Joint Undertaking (EuroHPC Joint Undertaking, EXT-2023E01-013) for the GPU hours.

\bibliography{bibliography}

\newpage

\appendix

\section{Prompts}\label{sec:prompts}

In this appendix, we provide the specific prompts used to interact with the the Large Language Models of this work. 

\begin{figure}[!h]
\centering
\includegraphics[scale=.65]{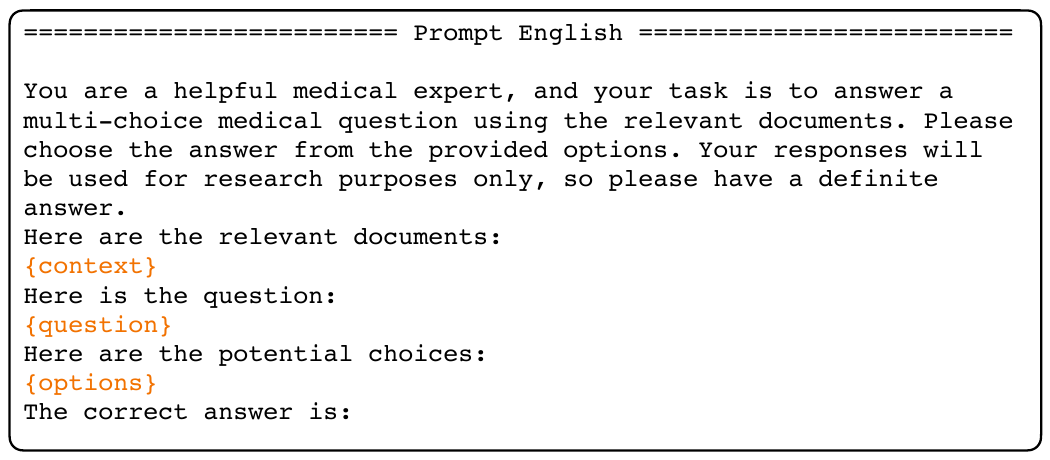}
\caption{Prompt used for models in English.}
\label{fig:pixt3_overview}
\end{figure}

\begin{figure}[!h]
\centering
\includegraphics[scale=.65]{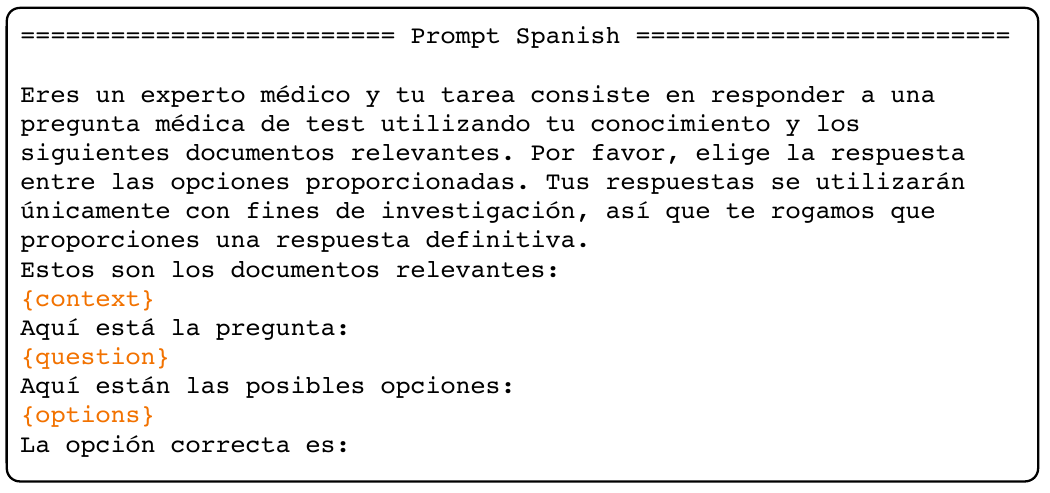}
\caption{Prompt used for models in Spanish.}
\label{fig:pixt3_overview}
\end{figure}

\begin{figure}[!h]
\centering
\includegraphics[scale=.65]{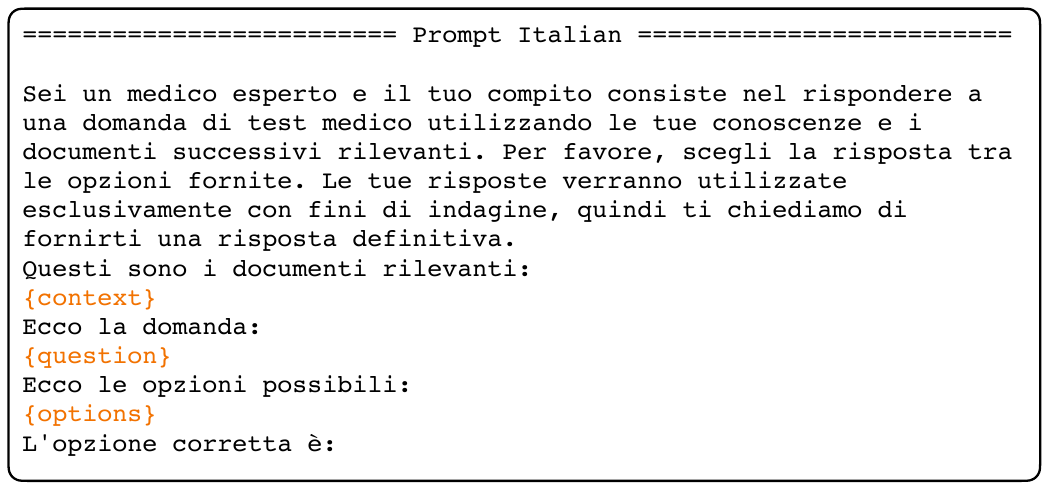}
\caption{Prompt used for models in Italian.}
\label{fig:pixt3_overview}
\end{figure}

\begin{figure}[!h]
\centering
\includegraphics[scale=.65]{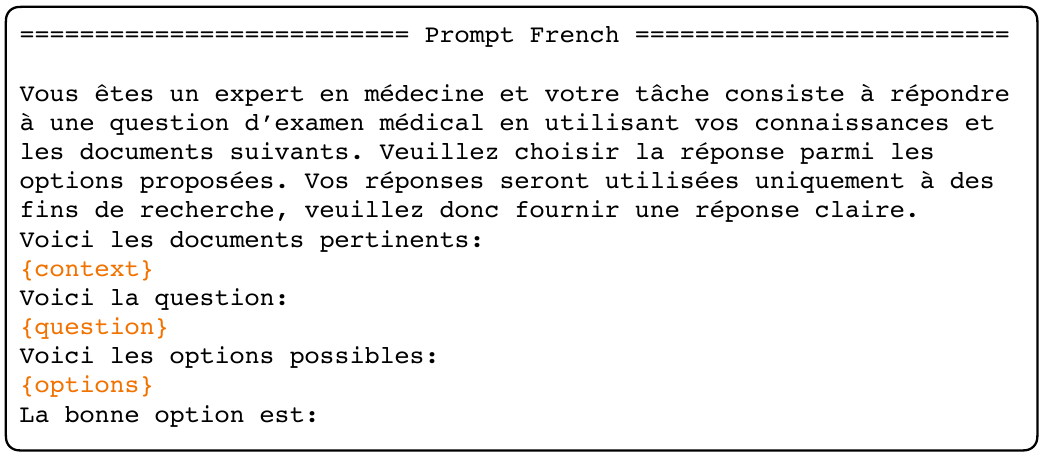}
\caption{Prompt used for models in French.}
\label{fig:pixt3_overview}
\end{figure}

\begin{figure}[!h]
\centering
\includegraphics[scale=.65]{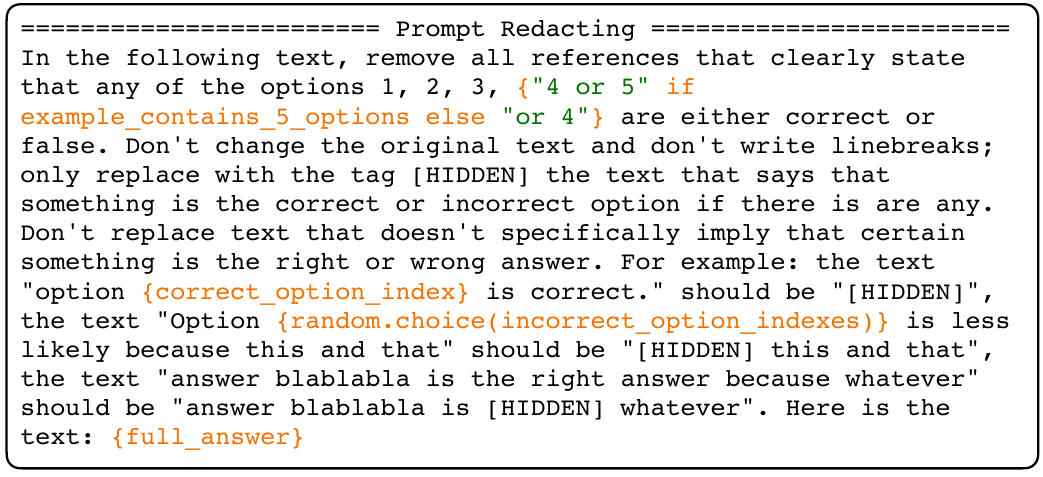}
\caption{Prompts to remove explicit references to the multiple-choice options.}
\label{fig:pixt3_overview}
\end{figure}

\pagebreak

\section{Hyperparameters}
\label{sec:hyperparameters}

In this appendix we list some of the hyperparameters used in this work.







\begin{table*}[!h]
\centering
\scriptsize
\begin{tabular}{p{7cm}p{3cm}}
\toprule
\textbf{Hyperparameter} & \textbf{Value} \\
\midrule
Optimizer &  adamw\_torch\_fused\\
Learning rate  & 0.00015 \\
Weight decay  & 0.0\\
ADAM $\epsilon$ & 1e-7\\
Epochs  & 10 \\
Train batch size  & 16\\
Evaluation batch size  & 8 \\
Floating Point 16-bit precision training & False\\
Brain Float 16-bit precision training & True\\
\midrule
Maximum \#tokens in input  &\\
\midrule
PMCLLaMA &  2048 \\
LLaMA2 &  4096\\
Mistral &  8000\\
BioMistral  & 8000\\ 
\midrule
Maximum \#tokens in generation  &\\
\midrule
PMCLLaMA &  2048 \\
LLaMA2 &  4146\\
Mistral &  8050\\
BioMistral & 8050\\ 
\midrule
Low-Rank Adaptation (LoRA) & \\
\midrule
R parameter & 8 \\
LoRA $\alpha$ &  16 \\
LoRA Dropout& 0.05\\
\bottomrule
\end{tabular}
\caption{Hyperparameters used in the configuration of the experiments.}
\end{table*}

\clearpage

\section{Efficiency metrics}
\label{sec:efficiency}

\textcolor{black}{In this work we only use or apply the LLMs to establish our benchmark, be that in zero-shot or fine-tuning. As such, we do not perform any modification in the way the LLMs work. Therefore, for efficiency and architectural issues the original papers of Llama2, PMC-Llama, Mistral and BioMistral could be inspected. 
Our contributions are focused on (i) establishing a multilingual benchmark for Medical QA, (ii) experimenting with state-of-the-art RAG methods and (iii) providing gold reference explanations as a form of "gold" RAG that can be used to compare the LLMs with. Having said that, below we offer detailed information about some efficiency metrics. All the metrics have been calculated using a NVIDIA A100 Graphics Processing Unit (GPU).
}

\begin{itemize}
    \item \textcolor{black}{The total number of parameters updated through Low Rank Adaptation (LoRA) during   Parameter-Efficient Fine-Tuning (PEFT) are the following: }
    
\begin{table*}[!h]
\centering
\scriptsize
\begin{tabular}{lccc}
\toprule
&\multicolumn{3}{c}{\textbf{7B parameter models}} \\
\midrule
&Trainable parameters & All parameters & Trainable \% \\ 
\midrule
Mistral and BioMistral &20,971,520 & 3,773,042,688 & 0.555825 \\
\midrule
&\multicolumn{3}{c}{\textbf{13B parameter models }} \\
\midrule
&Trainable parameters & All parameters & Trainable \% \\ 
\midrule
LLaMA2 & 31,293,440 & 6,703,272,960 &0.466838 \\
PMC-LLaMa & 31,293,440 & 6,703,283,200 &0.466838 \\
\bottomrule
\end{tabular}
\caption{Trainable parameters: Number of parameter in training using the LoRA model; All parameters: total of parameters used in the LoRA model; Trainable \%: number of trainable parameters of the total number of parameters in the LoRA model. }
\end{table*}
    
\item \textcolor{black}{Table \ref{tab:samplesSecond} shows the number of \textbf{samples per second} processed when using Mistral (7B) and LLaMA2 (13B) in a NVIDIA A100 GPU. The performance in the other two models, BioMistral (7B) and PMC-LLaMA (13B) is the same.}

\item \textcolor{black}{Table \ref{tab:times} shows \textbf{the time in minutes and hours} when processing data with Mistral (7B) and LLaMA2 (13B). The other two models, BioMistral (7B) and PMC-LLaMA (13B), showcase the same times.}

\end{itemize}

\begin{table*}[!h]
\centering
\small
\begin{tabular}{l|cc|cc}
\toprule
\textbf{Samples per second} & \multicolumn{2}{c|}{\textbf{Train}} & \multicolumn{2}{c}{\textbf{Inference}}  \\
\midrule
& 7B & 13B &7B & 13B \\
\midrule
E & 1.981  & 1.270 &7.681 &4.757 \\
H &1.998 & 1.282&7.676&4.76 \\
\midrule
None  & 3.248 &2.116 & 11.375 &6.956 \\
RAG-7  & 1.031 &0.629 &3.637 &2.081 \\
RAG-32& 0.191 & 0.281 &0.744 &1.013 \\
\bottomrule
\end{tabular}
\caption{Samples processed by second in a NIVIDIA A100 GPU. E: Full gold explanation. H: Full gold explanation with Hidden explicit references to the correct/incorrect answer; None: model without any additional external knowledge; RAG-7: Retrieval Augmented Generation with k=7; RAG-32: Retrieval Augmented Generation with k=32;}
\label{tab:samplesSecond}
\end{table*}

\begin{table*}[!h]
\centering
\small
\begin{tabular}{lrr}
\toprule
\textbf{Time for training} & 7B & 13B \\
\midrule
E & 1h 4m & 2h 1m \\
H & 1h 9m & 2h 9m\\
\midrule
None  & 47m & 1h 39m\\
RAG-7  & 1h 42m & 3h 2m\\
RAG-32& 7h 34m &5h 31m \\
\bottomrule
\end{tabular}
\caption{Time in minutes (m) and hours (h) when processing data in a NIVIDIA A100 GPU. E: Full gold explanation. H: Full gold explanation with Hidden explicit references to the correct/incorrect answer; None: model without any additional external knowledge; RAG-7: Retrieval Augmented Generation with k=7; RAG-32: Retrieval Augmented Generation with k=32.}
\label{tab:times}
\end{table*}

\end{document}